\documentclass[11pt]{article}

\usepackage[]{acl2023}

\usepackage{times}
\usepackage{latexsym}
\usepackage{tikz}
\usepackage{xcolor}
\usepackage{graphicx}
\usepackage{color}

\usepackage{enumitem}
\usepackage{graphicx}
\usepackage{balance}  
\usepackage{amsmath}
\usepackage[linesnumbered,ruled,vlined]{algorithm2e}
\usepackage{algpseudocode}
\usepackage{latexsym}
\usepackage{multirow}
\usepackage{multicol}
\usepackage{color}
\usepackage{dashrule}
\usepackage{extarrows}
\usepackage{dsfont}
\usepackage{centernot}
\usepackage{hyperref}
\usepackage{color}

\usepackage[T1]{fontenc}

\usepackage[utf8]{inputenc}

\usepackage{microtype}
\usepackage{graphicx}
\usepackage{subfigure}
\usepackage{booktabs} 

\usepackage{amsmath}
\usepackage{amssymb}
\usepackage{mathtools}
\usepackage{amsthm}
\usepackage{microtype}
\definecolor{nav}{RGB}{0,0,128}

\newcommand{\answerTODO}[1][]{\textcolor{red}{\bf [TODO]}}

\usepackage{booktabs}

\title{Pretraining Without Attention}
 \author{Junxiong Wang \\ Cornell \And  Jing Nathan Yan \\ Cornell  \And Albert Gu \\ DeepMind \And Alexander M. Rush \\ Cornell }

\begin{document}
\maketitle

\begin{abstract}
 Transformers have been essential to pretraining success in NLP. While other architectures have been used, downstream accuracy is either significantly worse, or requires attention layers to match standard benchmarks such as GLUE. This work explores pretraining without attention by using recent advances in sequence routing based on state-space models (SSMs). Our proposed model, Bidirectional Gated SSM (BiGS), combines SSM layers with a multiplicative gating architecture that has been effective in simplified sequence modeling architectures.
 The model learns static layers that do not consider pair-wise interactions. 
 Even so, BiGS is able to match BERT pretraining accuracy on GLUE and can be extended to long-form pretraining of 4096 tokens without approximation. Analysis shows that while the models have similar average accuracy, the approach has different inductive biases than BERT in terms of interactions and syntactic representations. 




\end{abstract}

\section{Introduction}




%
%

%



Transformers are the \emph{de facto} model architecture for NLP pretraining~\cite{vaswani2017attention}. 
Since BERT~\cite{devlin2018bert}, they have proven central to
NLP tasks with their ability to learn effectively on large unlabeled datasets. Specifically, the use of attention as a central routing component seems to be critical to empirical success on downstream tasks. Other architectures have been proposed but require attention layers for high-accuracy~\cite{tay2020efficient,lee2021fnet}.


Is the centrality of attention in pretraining due to inductive bias or computational convenience? This question is complicated by the properties of common sequence routing layers: recurrent neural network (RNN) models due not scale as well as attention, whereas convolutional neural networks (CNNs) can not easily model long-distance dependencies. 

State-space models (SSMs) for deep learning provide a promising alternative. Recent works show that SSMs 
are a competitive architecture for long-range sequence modeling~\citep{gu2021efficiently}. SSMs achieve strong results on speech generation~\cite{goel2022s} and on the Long Range Arena benchmark \citep{tay2020long} outperform standard and long-range transformer architectures~\citep{gu2021efficiently, gupta2022diagonal,gu2022parameterization, smith2022simplified}. In addition to improving accuracy, SSM-based routing does not have quadratic complexity as the length of the sequence grows.
Concretely, the model provides a way to achieve RNN-like long-range dependencies with CNN-like training speed. 


This work proposes an architecture for applying SSMs using a \emph{Bidirectional Gated SSM} (BiGS) model for BERT-style pretraining. BiGS uses SSM-routing at its core as a replacement for attention. However, this change alone significantly degrades the representational capacity of the model. To target this issue, we develop a multiplicative gating architecture  ~\cite{dauphin2017language, hua2022transformer,mehta2022long}. In combination, this leads to a simpler routing approach that remains surprisingly effective at modeling necessary interactions.



Experiments compare SSMs to standard NLP pretraining. While we find that SSMs by themselves underperform on NLP pretraining tasks, 
  BiGS is able to match the performance of a BERT model when trained on the same data in a controlled setting.  By additionally pretraining on longer-length instances, the model is able to grow without approximation to extend to input sequences of length 4,096. 
  Analysis shows that importance of multiplicative gating in fixing specific issues of variable-length textual input. All models from this work are available at \url{https://github.com/jxiw/BiGS}.

\section{Related Work}

Prior to BERT, promising pretraining approaches for learning contextual representations were learned using RNN-based models~\cite{mccann2017learned,DBLP:conf/naacl/PetersNIGCLZ18}.  While important precursors, their accuracy did not scale with data or compute as well as Transformers. This gap remains even when back-porting best-practices from Transformer pretraining~\citep{peters2019tune}. 
Recently \citet{tay2021pre} explored pretraining with several convolutional (CNN) variants. Results show that CNN without attention does not perform well, although they note benefits in routing speed. \citet{lee2021fnet} propose FNet which replaces the attention layer with a Fourier transform. Without attention, this achieves 92-97\% results on GLUE~\citep{wang2018glue}. 
Other works have used CNN-based models with multiplicative gating for NLP tasks such as machine translation~\citep{dauphin2017language}.
We believe BiGS is the first model to achieve BERT-level transfer learning on the GLUE benchmark without attention. 

Researchers have begun to use state-space models for NLP tasks, and have primarily focused on auto-regressive language modeling. In S4~\citep{gu2021efficiently} and its variants~\citep{gupta2022diagonal, gu2022parameterization}, researchers experimented with language modeling, achieving promising results, though slightly worse than transformers. Gated State Space adapts a SSM plus gating approach to language modeling~\cite{mehta2022long}. Concurrent to this work, \citet{dao2022hungry} propose H3 which closes the gap in auto-regressive language modeling, and with two attention layers outperforms transformers on OpenWebText. Finally, a related method, MEGA~\citep{ma2022mega} combines exponential moving average routing with a simple attention unit to outperform transformer baselines. Our approach instead focuses on bidirectional masked language modeling and questions of downstream generalization.

\section{Background}
\begin{figure}
    \centering
    \includegraphics[width=0.9\linewidth]{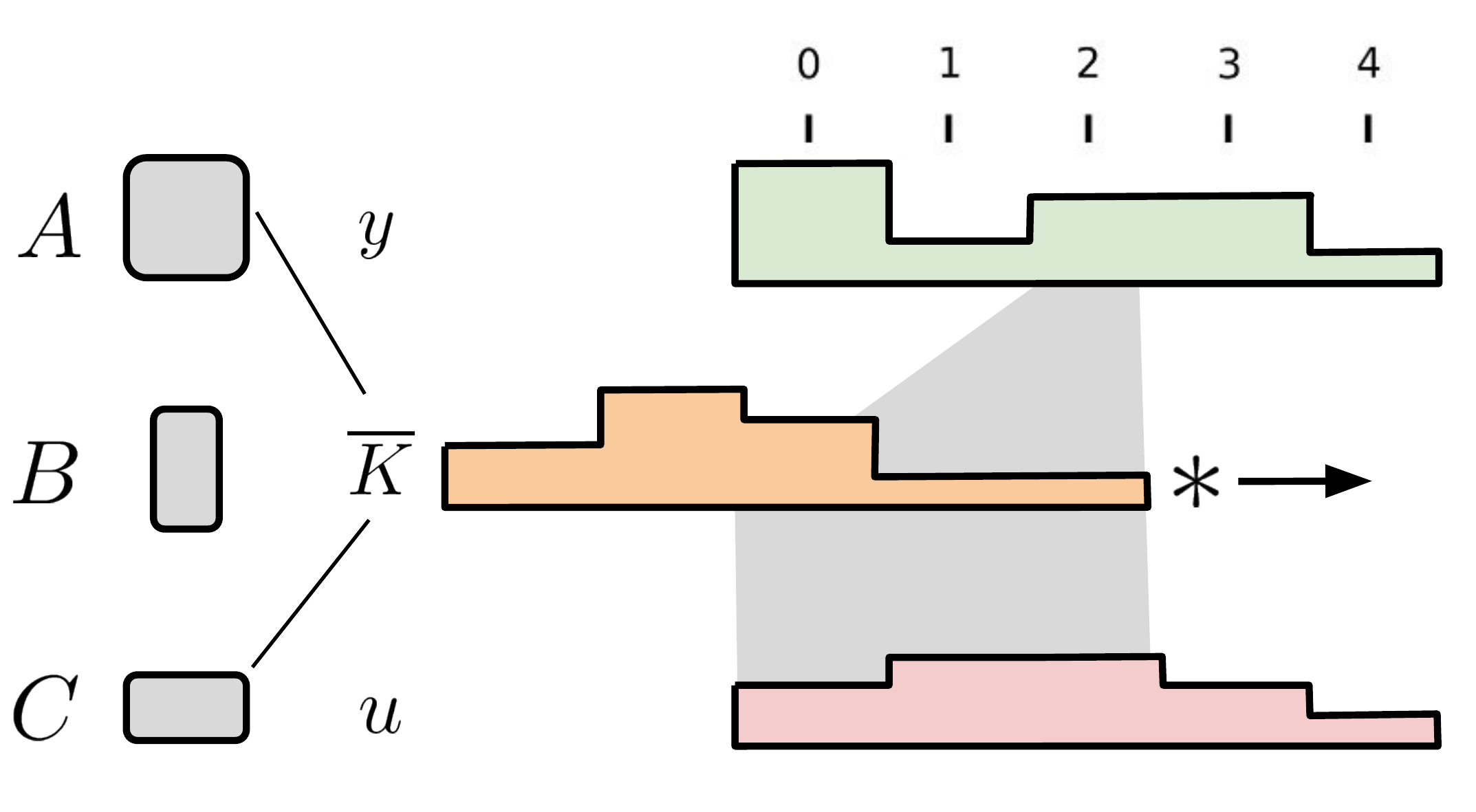}
    \caption{A SSM learns a one-dimensional kernel $\mathbf{\overline{K}}$, which is convolved with the input sequence $u$ to produce output $y$. Unlike attention, routing is static and does not depend on the input. In BiGS we use only two kernels per layer (forward and backward). Figure~\ref{fig:model:kernels} shows all the kernels used in the fully trained model.}
    \label{fig:ssm}
\end{figure}

\begin{figure*}[ht]
    \centering
    \includegraphics[width=1\linewidth]{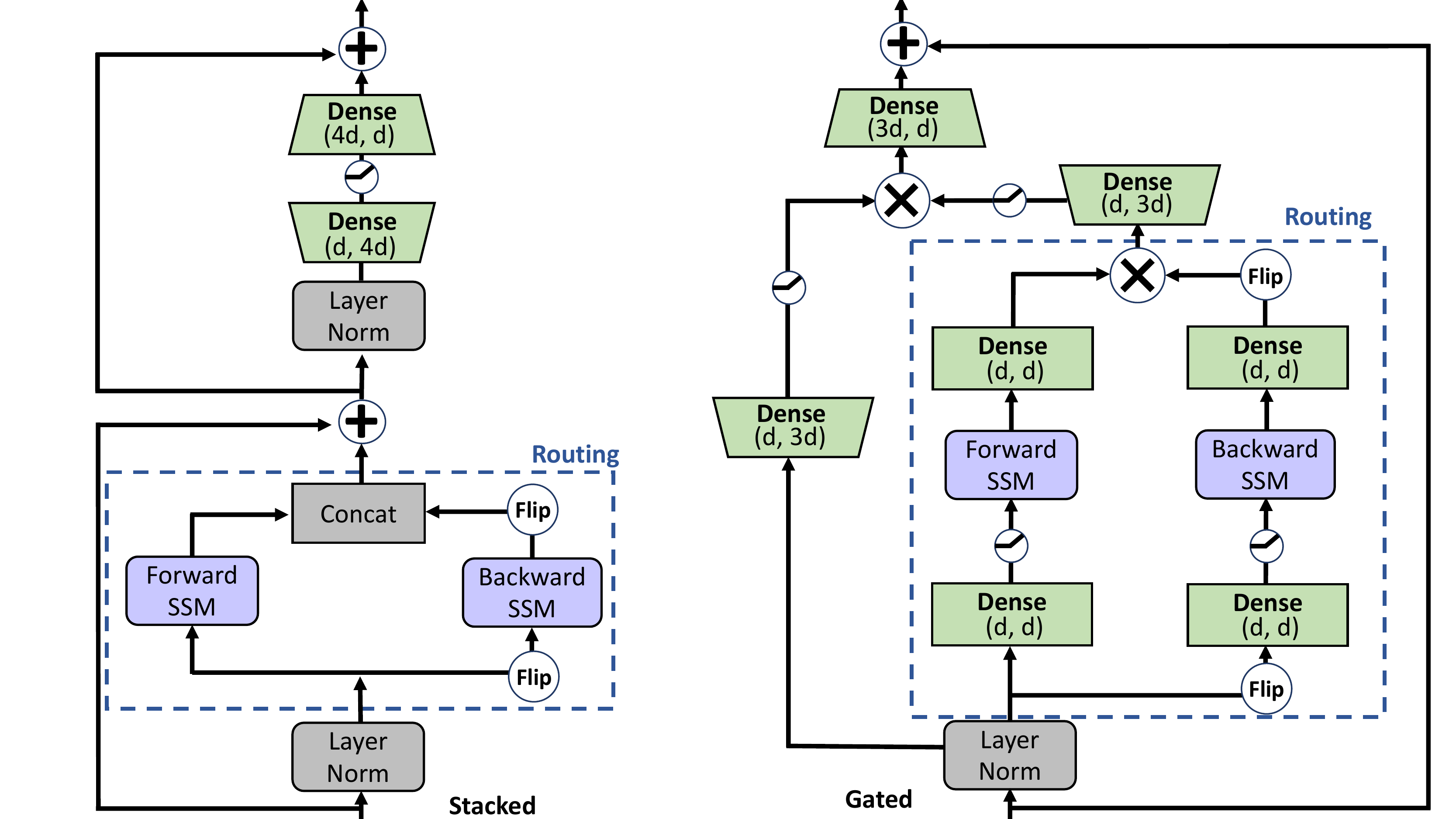}
    \caption{Model Variants.
     (\textsc{stack}) is the standard transformer architecture, (\textsc{gated}) is based on the gated unit~\cite{mehta2022long, hua2022transformer}. For the Routing component (dashed lines), we consider both a bidirectional SSM (shown) and standard self-attention. 
    The gate (\textbf{$\otimes$}) represents element-wise multiplication. The BiGS model uses \textsc{gated} with SSM. }\label{fig:model:architectures}
\end{figure*}

\subsection{State Space Models}

A state space model (SSM) is a general-purpose tool for describing the relationship between a continuous-time scalar input $u(t)$ to scalar output $y(t)$ by the following differential equations:
\begin{align*}
    x'(t) = \boldsymbol{A}x(t) + \boldsymbol{B}u(t),\ \ \ 
    y(t) = \boldsymbol{C}x(t) + \boldsymbol{D}u(t).
\end{align*}
\noindent Where $x(t) \in \mathbb{R}^{N}$ is a continuous-time state vector, $x'(t)$ is its derivative, and the equation is parameterized by $\boldsymbol{A} \in \mathbb{R}^{N\times N}, \boldsymbol{B}\in \mathbb{R}^{N \times 1}, \boldsymbol{C} \in \mathbb{R}^{1 \times N}, \boldsymbol{D} \in \mathbb{R}^{1\times 1}$.

When applied to a discrete-time scalar input sequence $u_1, \ldots u_L$, the SSM equations and parameters can be discretized, leading to the following recursion,
\begin{align*}
x_{k} = \boldsymbol{\overline{A}} x_{k-1} + \boldsymbol{\overline{B}} u_k,  \ \ \ 
y_k = \boldsymbol{\overline{C}} x_k + \boldsymbol{\overline{D}}  u_k. 
\end{align*}
Where $\boldsymbol{\overline{A}},\boldsymbol{\overline{B}},\boldsymbol{\overline{C}},\boldsymbol{\overline{D}}$ are functions of the original parameters and a discretization rate.

This equation can be computed like an RNN where $x_{k} \in \mathbb{R}^{N}$ is a hidden state at time $k$. Unlike an RNN though, the linearity of the recursion allows $y_1 \ldots y_L$ to be computed directly using a convolution with precomputed kernel $\boldsymbol{\overline{K}} \in \mathbb{R}^{L}$ , 
\begin{align*}
& \boldsymbol{\overline{K}} = (\boldsymbol{\overline{C}}\boldsymbol{\overline{B}}, \boldsymbol{\overline{C}}\boldsymbol{\overline{A}}\boldsymbol{\overline{B}}, \dots, \boldsymbol{\overline{C}}\boldsymbol{\overline{A}}^{L-1}\boldsymbol{\overline{B}}) \\
&y = \boldsymbol{\overline{K}} \ast u
\end{align*}
\noindent The process is illustrated in Figure~\ref{fig:ssm}. In a practical sense, after training, this kernel $\boldsymbol{\overline{K}}$ fully characterizes the SSM, i.e. the model is a 1D convolution with a very long kernel.

\subsection{Learning SSMs}

\citet{gu2020hippo, gu2021efficiently} demonstrate an effective  
approach for using SSMs in neural networks. The core insight is to propose a parameterization 
of the transition matrix $\boldsymbol{A}$, known as HiPPO,
\begin{align*}  
\boldsymbol{A}_{nk}= -
\begin{cases} 
(2n+1)^{1/2}(2k+1)^{1/2} & \text{if } n > k \\ n+1 &\text{if } n=k \\ 0 & \text{if } n < k 
\end{cases} 
\end{align*}
This matrix yields a stable training regime that can also be efficiently trained. 
The full model, S4, retains the SSM ability to model long-term sequences while being more efficient than RNNs to train. 

Recently, researchers~\cite{gu2022parameterization, gupta2022diagonal} have proposed simplified diagonalized versions of S4, which achieve comparable results with a simpler approximation of the original parameterization. In preliminary experiments, we used several different S4 parameterizations but did not find a significant difference in accuracy. Throughout the work, we use S4D as the parameterization.


While the specifics of SSM discretization, parameterizations, and training are beyond the scope of this work,
at a high-level, we note that each of the models leads to the form in the previous section. 
The model can be therefore be trained by backpropagation though the convolution and discretization without the serial bottleneck of RNNs, and applied without the quadratic cost of attention. Each SSM itself has $O(N^2) $ parameters, and we use $N=64$ throughout the work. 






\subsection{Multiplicative Gating}
Gating units have been widely used to improve the performance of various architectures such as MLP, CNN, and Transformers~\citep{dauphin2017language, shazeer2020glu, narang2021transformer}.  One example of such a gating unit is the Gated Linear Unit (GLU) which has been used effectively for CNN-based NLP systems ~\cite{dauphin2017language}.
Let $\mathbf{u}$ represent an input activation. GLU first computes both a gating vector and a linear transform, $\sigma(\mathbf{W} \mathbf{u})$ and $\mathbf{V} \mathbf{u}$ respectively. The output of the layer is then the element-wise product 
$\sigma(\mathbf{W} \mathbf{u}) \otimes (\mathbf{V} \mathbf{u})$.

Recent work has shown that gating can increase the performance of models using simplified routing. \citet{hua2022transformer} show that linear time attention models can benefit form improved gating. \citet{mehta2022long} propose a Gated State Space architecture using gating for unidirectional SSM models. Multiplicative gating may restore some of the interaction capacity from full attention-based interactions.




\section{BiGS Model}

We consider two different architectures for SSM pretraining: a stacked architecture (\textsc{stack}) and a multiplicative gated architecture (\textsc{gated}) shown in Figure~\ref{fig:model:architectures}. 

\paragraph{Transformer Architecture }

The \textsc{stack} architecture with self-attention is equivalent to the BERT / transformer model. We replace the attention block with two sequential SSM blocks to mimic the nature of bi-directional self-attention. 

\paragraph{Gated Architecture}
The \textsc{gated} architecture is a bidirectional adaptation of the gated unit of \citet{hua2022transformer}. Specifically, let $\mathbf{X}_i \in \mathbb{R}^{L\times d}$ be activations at the $i$-th layer where the length is $L$, and the model size is $d$. We use the activation GELU~\cite{hendrycks2016gaussian} for $\sigma$. The first stage computes, 

\vspace{-1em}
\begin{align*}
& \mathbf{X}=\text{LayerNorm}(\mathbf{X}_i )   && \in \mathbb{R}^{L\times d} \\
& \mathbf{V}=\sigma(\mathbf{W}_v \mathbf{X})     && \in \mathbb{R}^{L\times 3d} \\
& \mathbf{F}=\sigma(\mathbf{W}_{f} \mathbf{X})   && \in \mathbb{R}^{L\times d}  \\
& \mathbf{B}=\sigma(\mathbf{W}_{b} \text{Flip}(\mathbf{X})) && \in \mathbb{R}^{L\times d}
\end{align*}

\noindent The second stage uses 2 sequential blocks (i.e., a forward and backward SSM layer) with a multiplicative gate.

\vspace{-1em}
\begin{align*}
& \mathbf{U}_1 = \mathbf{W}_{u_1} \text{SSM}(\mathbf{F}) && \in \mathbb{R}^{L\times d}\\
& \mathbf{U}_2 = \mathbf{W}_{u_2} \text{SSM}(\mathbf{B}) && \in \mathbb{R}^{L\times d}\\
& \mathbf{U} = \sigma(\mathbf{W}_u (\mathbf{U}_1 \otimes \text{Flip}(\mathbf{U}_2))) && \in \mathbb{R}^{L\times 3d}
\end{align*}
\noindent The third stage uses a feed-forward layer again with gating, to replace the two dense blocks in the traditional transformer architecture. We sum this output $\mathbf{O}$ with the original input $\mathbf{X}_i$ finally as the input $\mathbf{X}_{i+1}$ of the next layer $i+1$. 

\vspace{-1em}
\begin{align*}
& \mathbf{O} = \mathbf{W}_o (\mathbf{U}\otimes \mathbf{V}) && \in \mathbb{R}^{L\times d},\\
& \mathbf{X}_{i+1} = \mathbf{O} + \mathbf{X}_i && \in \mathbb{R}^{L\times d}
\end{align*}

 







The number of parameters per layer in gated SSM is roughly $13d^2$ while the number of parameters per layer in the stack is $12d^2$. We compensate for this difference by using fewer gated layers.

\paragraph{SSM Layer}

The SSM layer under both architectures is a map over vector sequences, $\text{SSM}(\mathbf{X}):\mathbb{R}^{L\times d} \mapsto \mathbb{R}^{L\times d}$. However SSMs are defined for scalar sequences. Past work, creates $d$ differently parameterized SSMs for each dimension~\cite{gu2021efficiently}. Experimentally though, we found it just as effective to use the same parameterization (and therefore kernel $\mathbf{\overline{K}}$) for each hidden dimension. This simplifies model analysis and makes the total number of SSM parameters negligible.

\section{Experimental Setup}\label{sec:experiments}

Experiments compare the performance of SSM-based models to attention-based models on several standard fine-tuning benchmarks. Experiments control for total parameter-size and amount of pretraining in terms of number of tokens. All models are on the order of magnitude of BERT-Large at around 350M parameters; all \textsc{gated} SSM models use 23 layers and \textsc{stack} models 24 to match parameter count.
In order to run ablation tests, we consider three different pretraining scales: 11B (short), 29B (medium), and 97B (full). Models and architectures are roughly similar in training speed at this length.
The 11B (short) training scale is roughly equivalent to the "24h BERT" setting typically used in research studies~\cite{izsak2021train}. Full training is closer to the original BERT model which was trained on 128B tokens. 


For all pretraining, we follow the training data and masking strategy of \citet{izsak2021train}. Following RoBERTa~\cite{liu2019roberta}, we use only masked language modeling and not next-sentence prediction. We preprocess and mask tokens offline for all models for consistency, with maximal sequence length to be 128. 
We use a grid search on perplexity to select configurations of weight decay and learning rate; other hyperparameters follow \citet{izsak2021train}.  For SSM, we use a cosine decay learning rate scheduler, which starts at 0, warms up to the peak learning rate, and then decays back~\cite{gu2021efficiently}. 


\begin{table*}[t]
\centering 
\label{glue}
\begin{tabular}{lc|rrrrrrrr|r}
\toprule
               &  Arch / Route & MNLI & QNLI &  QQP & RTE  & SST2 & MRPC & COLA &  STS\textsubscript{B} & AVG \\
 &   & 393k & 105k & 364k & 2.5k &  67k  & 3.7k & 8.5k & 7k & \\ 
\midrule
&& \multicolumn{9}{c}{Short Training / $\sim 11$B Tokens }  \\
\midrule
BERT  & \textsc{stack} / \textsc{att} & 82.7 & 90.1 & 87.7 & 76.8 & 91.5 & 90.8 & 58.6 & 88.6 & 83.3 \\ 
      & \textsc{stack} / \textsc{ssm} & 78.4 & 83.5 & 85.6 & 60.5 & 91.6 & 83.9 & 53.1 & 81.3 & 77.2 \\ 
      & \textsc{gated} / \textsc{att} & 82.2 & 88.3 & 87.4 & 71.7 & 91.3 & 88.5 & 58.8 & 86.5 & 81.8 \\ 
BiGS  & \textsc{gated} / \textsc{ssm} & 82.6 & 89.2	& 87.6 & 73.8 & 92.8 & 88.9 & 63.2 & 88.4 & 83.3 \\
\midrule
&& \multicolumn{9}{c}{Medium Training / $\sim 29$B Tokens }\\
\midrule
BERT & \textsc{stack} / \textsc{att}  & 85.0 & 90.9 & 87.9 & 80.5 & 93.0 & 90.9 & 60.8 & 89.2 & 84.8 \\ 
  & \textsc{stack} / \textsc{ssm}     & 80.1 & 86.5 & 87.2 & 65.6 & 92.3 & 86.5 & 56.5 & 83.4 & 79.8 \\ 
  & \textsc{gated} / \textsc{att}     & 83.5 & 90.2 & 87.6 & 72.0	& 91.7 & 88.7 & 61.6 & 87.5	& 82.9 \\ 
BiGS  & \textsc{gated} / \textsc{ssm} & 84.5 & 90.2	& 88.3 & 78.6 & 94.4 & 89.6 & 63.9 & 89.3 & 84.8 \\ 
\midrule
&& \multicolumn{9}{c}{Full Training / $\sim 97$B Tokens }\\
\midrule
BiGS  & \textsc{gated} / \textsc{ssm} & 86.2 & 90.9	& 88.3 & 79.4 & 94.6 & 89.5 & 67.3 & 90.1 & 85.8 \\
\midrule
&& \multicolumn{9}{c}{Non-Attention Based Pretraining} \\
\midrule
CNN & \textsc{stack} /  \textsc{cnn} & $\sim$75 & - &  - & - & 92.2 & - & - & - & -\\
ELMo & \textsc{stack} / \textsc{rnn} & 68.6 & 71.2 & 84.3 & 53.4 & 91.5 & 70.5 & 44.1 & 82.3 & 68.7\\
FNet\textsubscript{L} & \textsc{stack} / \textsc{fnt} & 78.0 & 85.0 & 85.0 & 69.0 & 94.0 & 88.0 & - & 84.0 & - \\
\midrule
\midrule
&& \multicolumn{9}{c}{GLUE Test Result} \\
\midrule
\small{BERT\textsubscript{1}} & \textsc{stack} /  \textsc{ssm} & 86.7/85.9 & 92.7 & 72.1 & 70.1 & 94.9 & 88.9 & 60.5 & 86.5 & 79.6 \\ 
\small{BERT\textsubscript{2}} & \textsc{stack} /  \textsc{ssm} & 86.0/85.2 & 92.6 & 72.0 & 78.3 & 94.5	& 89.9 & 60.9 & 87.5 & 83.0 \\ 
BiGS  & \textsc{gated} / \textsc{ssm} & 86.1/85.0 & 91.6 & 71.2 & 77.6 & 94.9 & 88.7 & 64.4 & 87.5 & 83.0 \\ 
\bottomrule
\end{tabular}
\caption{GLUE Results. (Top) Comparison of different architectures and routing in a controlled setting~\cite{izsak2021train}. See Figure~\ref{fig:model:architectures} for details. We fine-tune RTE, MRPC, and STS-B from a MNLI checkpoint following the convention by ~\cite{izsak2021train}. We report accuracy by averaging the results of six runs for MNLI, QNLI, RTE, SST-2 and F1 score for QQP, MRPC and Matthew’s correlation for CoLA and Spearman’s correlation for STS-B. All models are comparable to BERT-Large in size. (Bottom) Reported comparable results for other non-attention-based pretraining models based on CNNs, LSTMs and FNet~\cite{DBLP:conf/naacl/PetersNIGCLZ18, tay2021pre,lee2021fnet,wang2018glue}. BERT\textsubscript{1} represents the official BERT result~\cite{devlin2018bert}, and BERT\textsubscript{2} represents the result using an MNLI checkpoint for other NLI tasks~\cite{izsak2021train}.
We use $-$ to denote those results were not reported by previous research. }\label{tab:glue}
\end{table*}

Pretraining is done with length 128 token sequences. In order to adapt to longer sequences we apply continued pretraining. To adapt to 512 tokens for the SQuAD dataset, we follow the protocol of \citet{wettig2022should} and train on longer sequences of the same pretraining dataset.  To adapt to 4,096 tokens, we follow the Longformer~\cite{beltagy2020longformer} protocol and continue training the BiGS model on the text of length up to 4,096 tokens long, for 10k more steps using their proposed training corpus of longer documents. For 4,096 tokens, use a smaller BiGS model (119M) so that it is comparable in size Longformer-base and BART-base models. We note that Longformer (LED) and BART are based on superior underlying models that are trained significantly longer.



Our SSM implementation is based on the Annotated S4\footnote{https://srush.github.io/annotated-s4} \citep{rush2022s4}, and our pretraining uses the template from Hugging Face Transformers\footnote{https://github.com/huggingface/transformers} \citep{wolf2020transformers}. We experimented with variants of S4 SSMs and found they performed similarly; experiments use S4D~\cite{gu2022parameterization} for simplicity. Note that for a fair comparison, we keep the size of gated architecture comparable to a stacked architecture and our BERT's implementation.

\section{Results}

\subsection{GLUE}

Table~\ref{tab:glue} (Top) shows the main results for different pretrained models on the GLUE benchmark. In short and medium training, we note that the \textsc{stack} architecture is significantly better with attention than with SSM-routing. 
However, with the \textsc{gated} architecture, the SSM achieves competitive results. To confirm this is not simply from a better architecture, we try gating with attention but find it does not improve. On full training, BiGS continues to improve in accuracy.

Table~\ref{tab:glue} (Bottom) compares the BiGS architecture to other reported results on GLUE. First, we compare to other non-attention based pretrained models based on RNNs and CNNs~\citep{peters2019tune,tay2021pre,lee2021fnet}. Results from these works all show significant degradation in transfer learning with GLUE scores far below BERT. Next, we compare BiGS to the full BERT results as reported in past work, both from the original paper~\citep{devlin2018bert} and from follow-up works with an improved fine-tuning convention~\citep{izsak2021train}. We see that the  BiGS model achieves comparable test scores. 
While the final GLUE score is nearly identical we do see that the models perform differently on the underlying tasks, which we explore more below.



We also apply BiGS to SQuAD~\citep{rajpurkar2016squad}. SQuAD requires extending the length of the model 
from 128 to 512 tokens through additional training. We report the F1 score in Table~\ref{tab:squad}. 
We see that BiGS outperform BERT when adapted with this procedure~\citep{wettig2022should}.
We note that both of these models underperform BERT which was pretrained fully on 512 tokens. 

\begin{table}[tb]
    \centering
    \begin{tabular}{ll|c}
    \toprule
           & & SQuAD 1.1 \\
    \midrule
         BERT & (512) & 90.9\\
         \midrule
         BERT &(128 $\rightarrow$ 512) & 87.3 \\
         BiGS &  (128 $\rightarrow$ 512) & 89.5 \\
    \bottomrule
    \end{tabular}
    \caption{SQuAD F1 Dev Results. Models are trained by adapting full 128 token 
        models to 512 tokens~\cite{wettig2022should}, which under-performs 512-token BERT~\citep{devlin2018bert}. }
    \label{tab:squad}
\end{table}

\subsection{Long-Form Classification}

\begin{table}[tb]
    \centering
    \begin{tabular}{lr|cc}
    \toprule
        & Length & QALT & CNLI \\
    \midrule
         LED  & 1024    &  26.6/27.2  & 73.4\\
         LED  & 4096    &  26.6/27.3  & 71.5\\
         LED  & 16384   &  25.8/25.4   & 71.5\\
         \midrule
         BART & 256  &  26.0/25.8 & 69.8\\
         BART & 512  &  26.8/27.4 & 71.6\\
         BART & 1024 &  26.0/25.9 & 77.4\\
         \midrule
         BiGS & 128  & 32.3/30.0 & 68.7 \\
         BiGS & 4096 & 32.8/31.7 & 71.4 \\
    \bottomrule
    \end{tabular}
    \caption{SCROLLS Encoder Test set results. Baseline models are both encoder-decoder models, one based on Longformer (LED)~\cite{beltagy2020longformer} and the other on BART~\cite{lewis2019bart}. Inputs are truncated at length.}
    \label{tab:scroll}
\end{table}

An advantage of SSM-based routing is that the models can extend to longer-ranges without requiring approximation. To adapt to longer range classification, we continue pretraining on longer data (4,096). Table~\ref{tab:scroll} shows results on encoder-supported\footnote{We have not attempted to develop a BiGS model that performs generation.} experiments in the SCROLLS~\cite{shaham2022scrolls}, a recent long-range language modeling benchmark. We can compare the model to Longformer Encoder-Decoder (LED) and BART. On these long-range tasks, it performs as well or better, taking advantage of the long-range context.

\section{Analysis}

\subsection{Role of SSM}

\begin{figure}[t]
    \centering
    \includegraphics[width=1\linewidth]{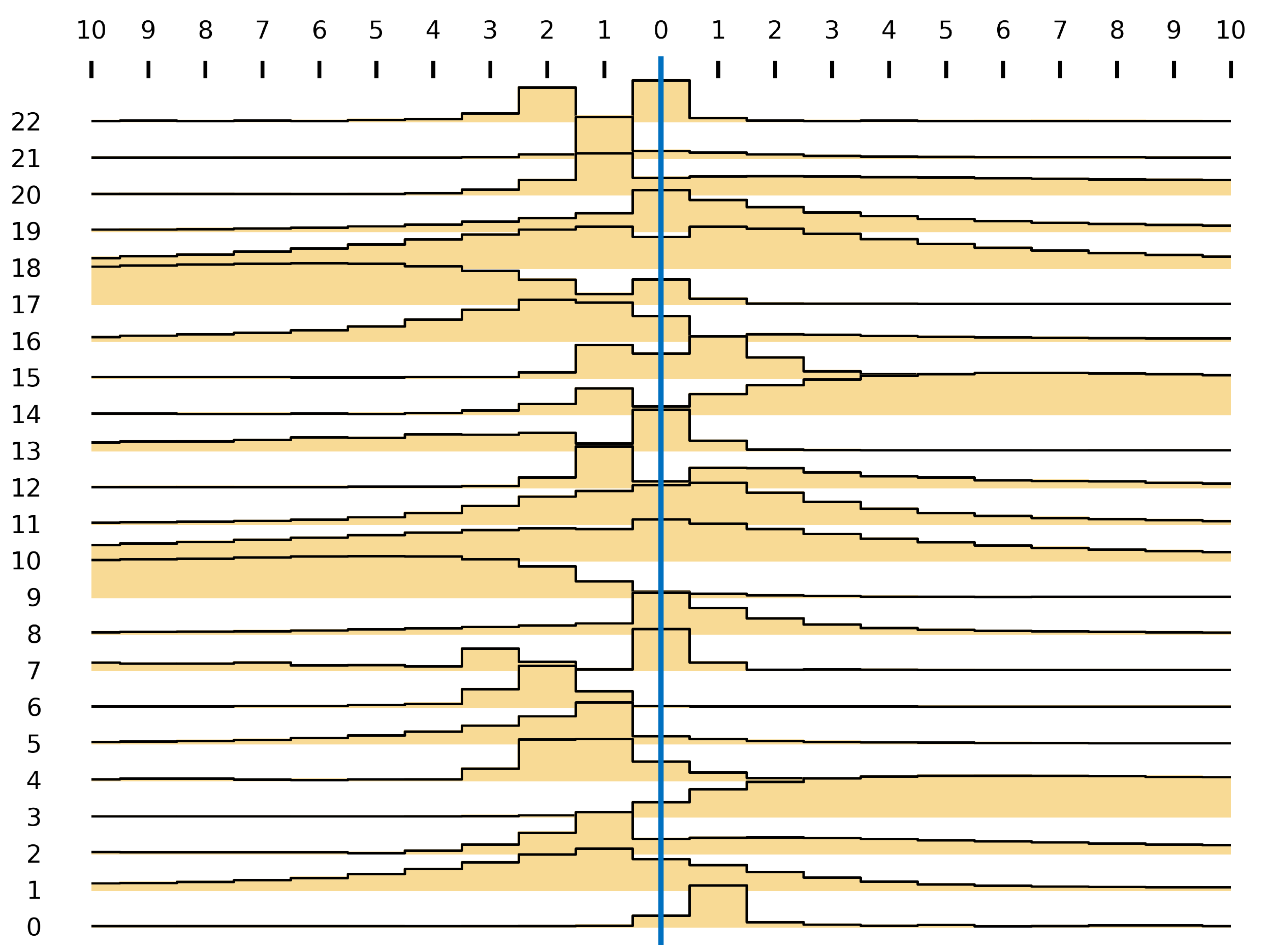}
    \caption{
    Complete SSM routing learned in BiGS. 
    Shows forward and backward kernels $\mathbf{\overline{K}}$ at each layer (0-22). 
    Values indicate absolute value of contribution of each relative position (-10, $\ldots$, 10) cropped from the full 2 $\times$ 128.
    Min-max scaling of absolute values is used for visual normalization.
    }\label{fig:model:kernels}

\end{figure}

\begin{figure}[t]
    \centering
    \includegraphics[width=1\linewidth]{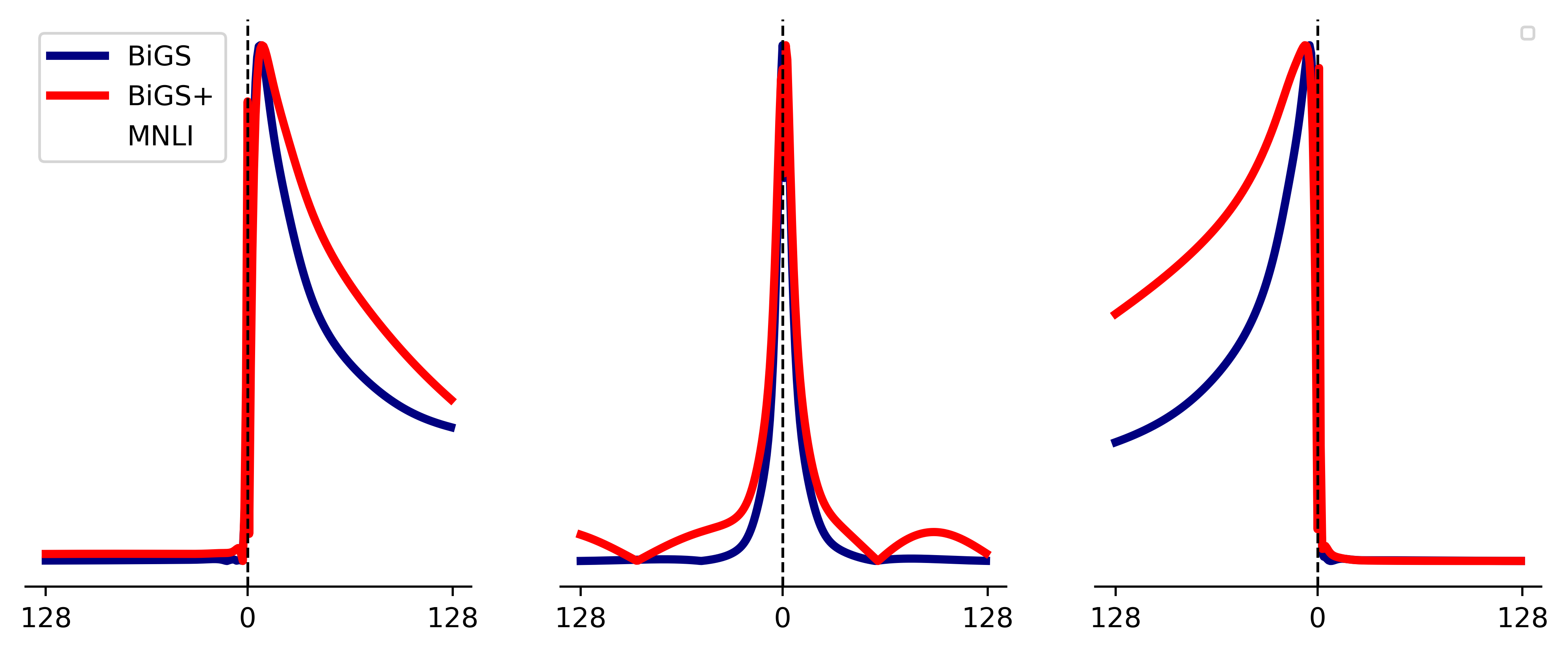}
    \caption{Change in SSM kernel after finetuning. 
    Shows $\mathbf{\overline{K}}$ after pretraining and after MNLI finetuning for Layer 14, Layer 18, and Layer 17 over all relative positions(-128, $\ldots$ , 128).  
    }\label{fig:kernels:comparison}
\end{figure}

Compared to multi-head attention where routing is determined by $L^2$ attention coefficients per head per layer, the BiGS SSM routing is relatively compact. Each layer has only $2L$ static values. Figure~\ref{fig:model:kernels} shows these values in the form of the forward and backward kernels. These kernels correspond partially to local aggregations such as the next word (layer 1) or a preceding trigram (layer 6), and partially to long-term future or past information (layer 14, layer 17). 

Figure~\ref{fig:kernels:comparison} shows how these kernels change during finetuning. In particular, during MNLI finetuning, the model needs to look at more long-distance information to match between sentences. This results in most local kernels remaining the same, but long distance kernels adjusting. The figure shows three kernels expanding their scope outward.

\subsection{Role of Gating}

\begin{figure}[ht]
    \centering
    \includegraphics[width=0.8\linewidth]{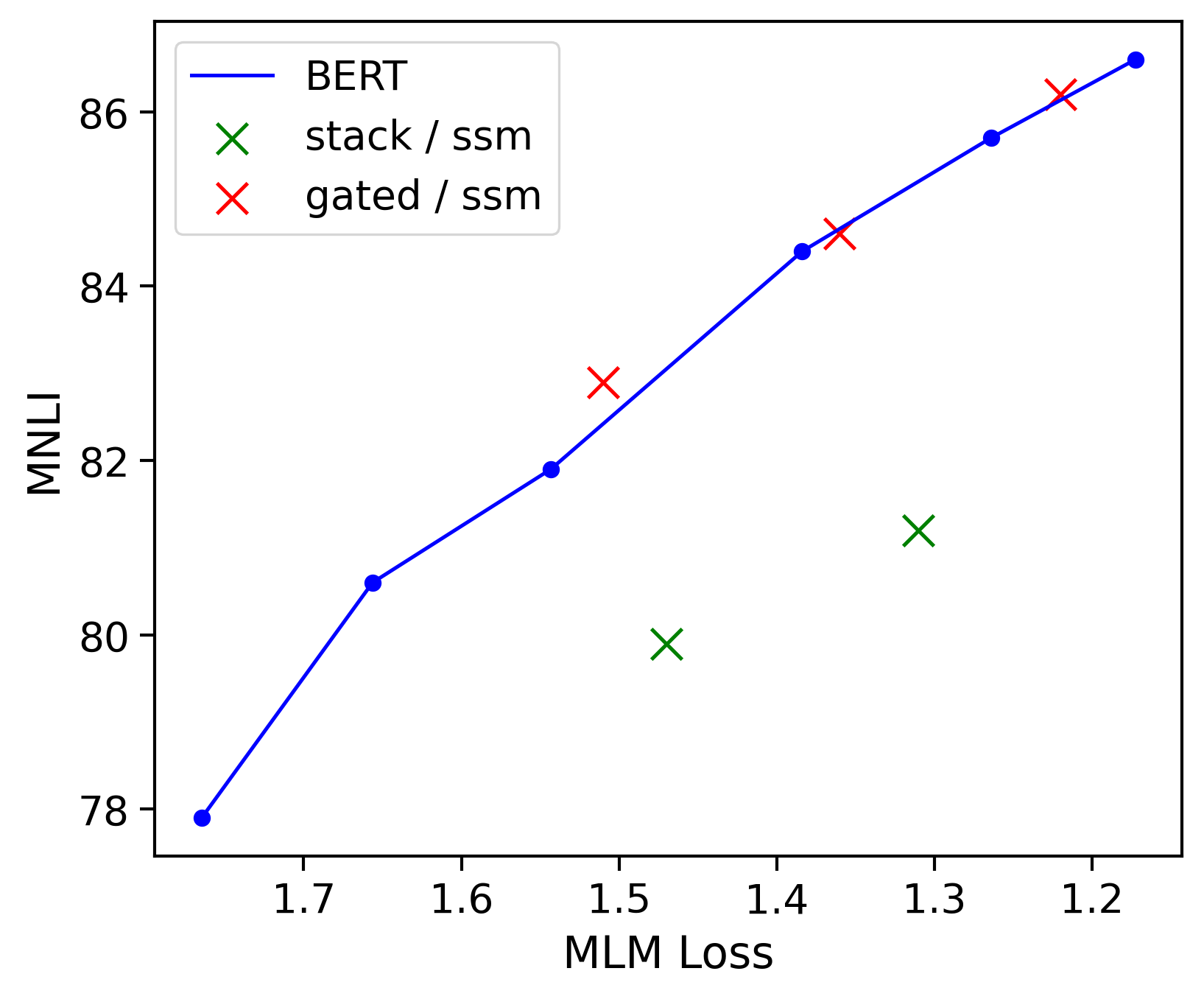}
    \caption{Comparison of pretraining loss and downstream accuracy. 
    Plots MNLI accuracy with respect to MLM loss.  BERT values from \cite{devlin2018bert}. Gated SSM shows similar transferability as BERT, whereas Stack SSM does not.}
    \label{tab:ppl}
\end{figure}

GLUE results show a significant improvement in downstream accuracy with the \textsc{gated} model; 
however, we find that \textsc{stack} SSM has a similar pretraining MLM loss.  Figure~\ref{tab:ppl} illustrates the difference of MLM loss and MNLI accuracy for both \textsc{gated} and \textsc{stack} SSM, compared to the MLM loss and expected MNLI values presented in BERT~\cite{devlin2018bert}. 
The figure shows that the gated model tracks closely the anticipated pretraining gains, while stack does not.


\begin{figure}[ht]\vspace{-1em}
\center
    \includegraphics[width=0.95\linewidth]{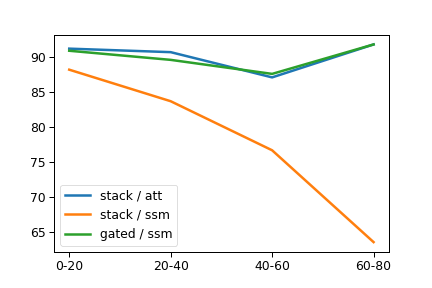}
    \caption{Accuracy by binned length on QNLI.}
    \label{tab:len}
\end{figure}

\begin{table}[t]
\center
    \begin{tabular}{lccc}
    \toprule
    \centering
    Model & Arch  & H P &   H $\sim$ P \\    
    \midrule
               & \textsc{stack} / \textsc{ssm} & 77.4 &  69.7\\
         BiGS  & \textsc{gated} / \textsc{ssm} & 77.4 &  77.7\\
    \bottomrule
    \end{tabular}
    \caption{Adversarial variant of QNLI. Distractor phrases ($\sim$) are added between the Hypothesis and Premise to increase the distance between relevant phrases. }
    \label{tab:synthetic}
\end{table}

We speculate that multiplicative gating helps the SSM model generalize to long-distance interactions. Table ~\ref{tab:len} compares accuracy of examples binned by length on the QNLI task. We see that the stack SSM decreases greatly in accuracy as the bins get larger, unlike for BERT and BiGS.  We further probe this issue with a synthetic adversarial task. We collect a dataset of examples from QNLI that have the same label prediction from both the \textsc{stack} and \textsc{gate} versions of the SSM model. We add distractor phrases ($\sim$), chosen to be non-ambiguous, in between the hypothesis (H) and premise (P) to see if the model has learned to skip over these to match H and P. Table~\ref{tab:synthetic} shows that the gated model performs significantly better at this task.

\begin{table}[t]
\centering
\resizebox{0.5\textwidth}{!}{
\begin{tabular}{lrrr}
\toprule
& BiGS & BERT & LSTM  \\
\midrule
\textsl{SUBJECT-VERB:}    &       &      &        \\
Simple                              & 100.0 & 100.0& 94.0    \\
Sentential complement          & 85.1  & 85.6 & 99.0   \\
Short VP coordination               & 91.0  & 86.5 & 90.0    \\
Long VP coordination                & 97.5  & 97.5 & 61.0    \\ 
Across prep phrase       & 88.6  & 84.8 & 57.0  \\ 
Across subj relative clause    & 88.4  & 84.9 & 56.0   \\
Across obj relative clause    & 89.9  & 85.1 & 50.0  \\
Across obj relative (-that) & 86.9  & 81.1 & 52.0  \\ 
In  obj relative clause        & 97.2  & 99.1 & 84.0  \\
In obj relative (-that)     & 88.7  & 81.6 & 71.0  \\
\midrule
\textsl{REFL ANAPHORA:}        &       &      &             \\
Simple                              & 97.1  & 98.9 & 83.0    \\
In a sentential complement          & 79.9  & 86.2 & 86.0   \\
Across a relative clause            & 79.1  & 75.9 & 55.0  \\
\bottomrule
\end{tabular}
}
\caption{Results on the \citep{marvin2018targeted} stimuli. BiGS and BERT results are comparable as they are in the exact same experiment setup. Numbers of LSTM models are taken from \citep{goldberg2019assessing}.}\label{tab:marvin_syntax}
\end{table}




\subsection{Syntactic Analysis}
\begin{figure}[ht]\vspace{-1em}
\center
    \includegraphics[width=0.95\linewidth]{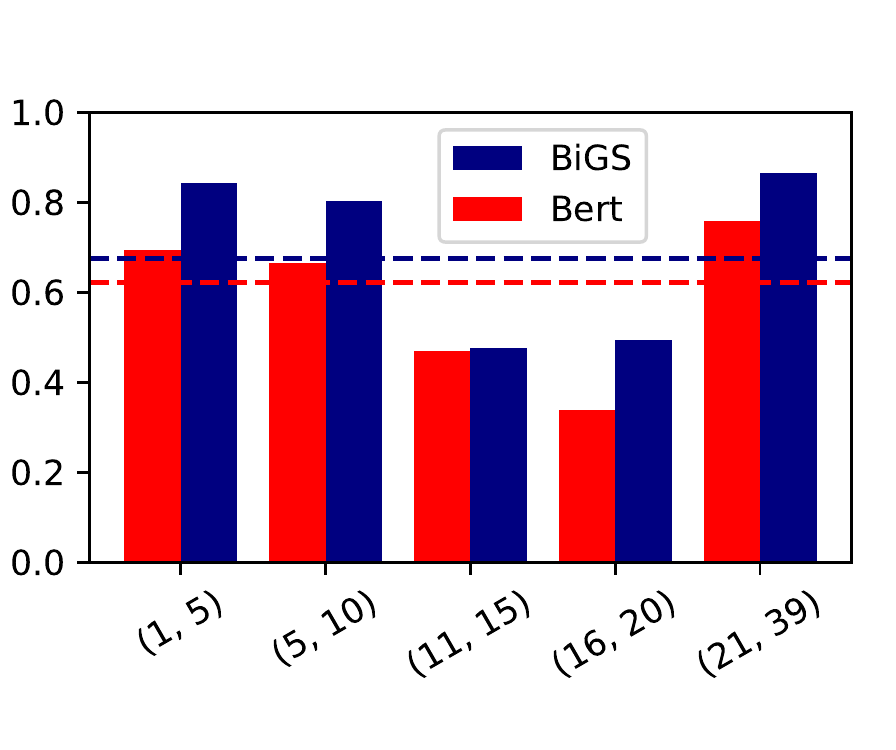}\vspace{-2em}
    \caption{Performance of CoLA w.r.t sentence length using matthews correlation coefficient(MCC). The red and navy dashed lines in the graph represent the mean value obtained from multiple rounds of evaluation.}
    \label{tab:cola:len}
\end{figure}

\begin{figure}
    \centering
    \includegraphics[width=\linewidth]{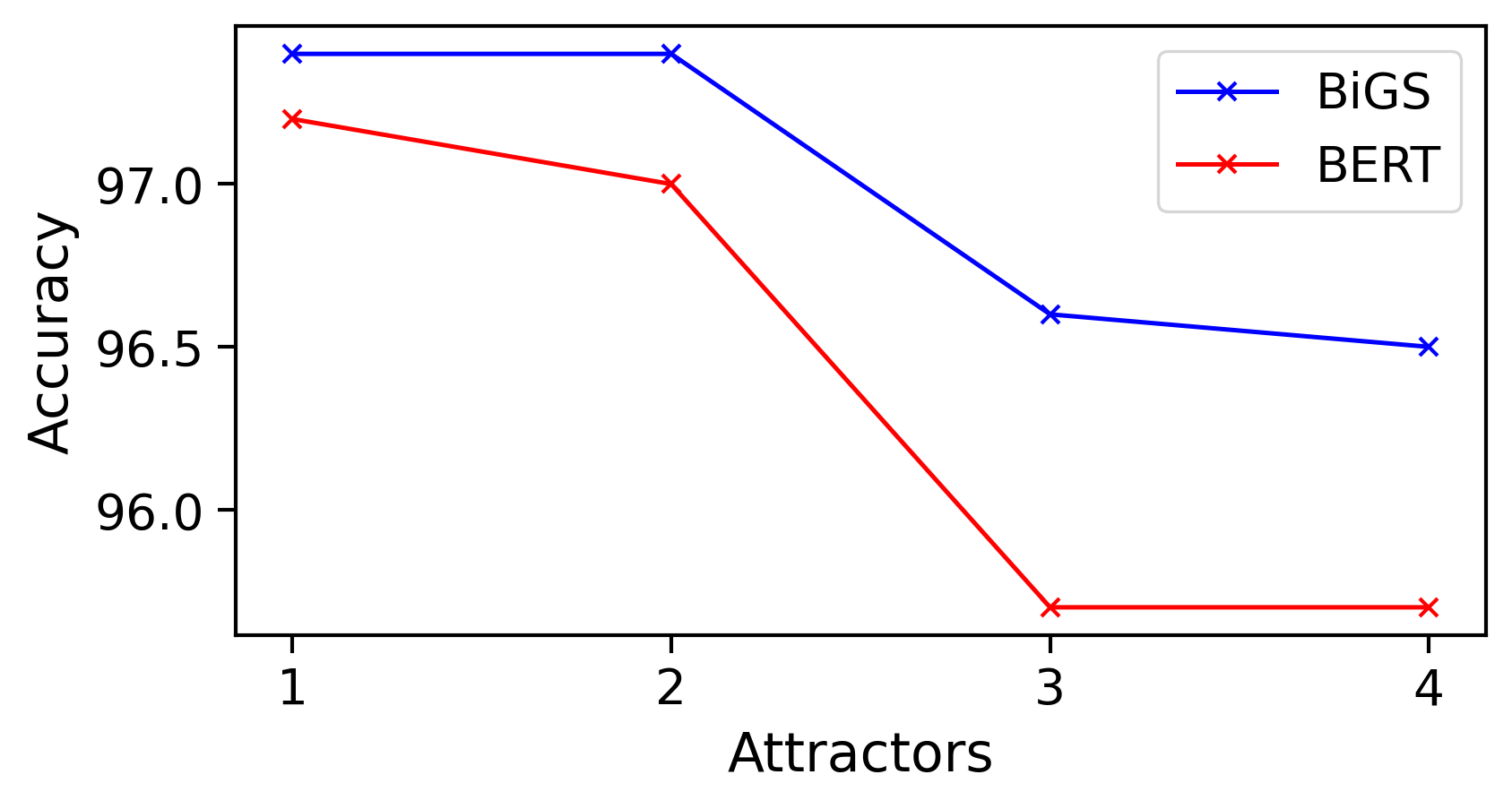}
    \caption{Syntactic Attractors task from \cite{linzen2016assessing}. Tests ability of models to match word agreement in the presence of intervening attractors. }
    \label{tab:linzen_syntax}
\end{figure}

BiGS seems to perform well on syntactic tasks such as CoLA~\cite{warstadt2019neural} (Figure~\ref{tab:cola:len}). We speculate that these results indicate that SSM-routing may have different inductive biases than transformer, in particular in terms of locality. We follow \citet{goldberg2019assessing} in adapting two preliminary experiments with of syntactic tests for masked language modeling:

\citet{linzen2016assessing} test a model's ability to distinguish agreement in the presence of spurious intervening "agreement attractors". For example the sentence "Yet the \textbf{ratio} of \underline{men} who survive to the \underline{women} and \underline{children} who survive [is] not clear in this story" has three attractors for the masked work [is].  Figure~\ref{tab:linzen_syntax} shows that BiGS consistently outperforms BERT as number of attractors grows. 

\citet{marvin2018targeted} develop pairs of manually constructed examples targeting various syntax phenomena and difficulties. Given a pair of examples from this stimuli: \emph{``\underline{No} students have ever lived here" } and \emph{``\underline{Most} students have ever lived here"}, we feed an adapted version \emph{``\underline{[MASK]} students have ever lived here"} into a model and compare the predicted scores for the masked position ``No'' and ``Most'' from it. Results are reported in Table~\ref{tab:marvin_syntax} and again show that SSM outperforms BERT on several agreement phenomena. 

While more experiments are needed, it is possible that the sequential nature of BiGS leads to an inductive bias to a more stack-like representation, since it cannot rely only on long-range matching. 










\subsection{Efficiency and FLOP Analysis}
Table ~\ref{tab:flops} gives the Floating Point Operations (FLOPs) for both BiGS and BERT models. FLOPs measures a best case computational cost of models. By comparing the FLOPs of BiGS and BERT for different input token lengths, we can better understand their relative efficiency and scalability. We calculate the training complexity, including both forward and backward passes for both BiGS and BERT, assuming single instance per batch. 
When the input token length is 128, BiGS shows slightly lower FLOPs than BERT, indicating a marginal advantage in terms of computational complexity. As the input token length increases to 512, BiGS surpasses BERT by a noticeable margin. This  increasing efficiency gap trend continues nonlinearly with token lengths of 1024 and 4096 respectively, implying that BiGS is better equipped to handle applications with longer input sequences. 

While SSMs have a theoretical computational benefit compared to attention, current implementations and hardware do not yet show this benefit. In our experiments, models are roughly the same speed. On longer range tasks, the SSM element is more tractable than the attention component, but concerns like training memory require careful optimization. Related work by~\citet{dao2022hungry} considers specialized kernels for SSM computation.

\begin{table}[tb]
    \centering
    \begin{tabular}{lcc}
    \toprule
        Length & BiGS & BERT \\
    \midrule
         128   & 8.1E+10 & 7.9E+10 \\
         512   & 3.2E+11 & 3.4E+11 \\
         1024  & 6.5E+11 & 7.2E+11 \\
         4096  & 2.6E+12 & 4.1E+12 \\
    \bottomrule
    \end{tabular}
    \caption{FLOP comparison between BiGS and BERT with respect to input token length. We calculated FLOP with a batch size of 1 and considered both the forward and backward passes.}
    \label{tab:flops}
\end{table}

\section{Limitations}

While SSMs are a promising technology for pretraining, there are some limitations for their use.

\begin{itemize}
    \item This work considers SSMs for bidirectional pretraining, and not autoregressive language modeling. In this setting, some of the benefits of SSMs are less apparent, such as being able to utilize RNN based generation. 

    \item In our preliminary studies in applying BiGS to long-range question answering (WikiQA~\cite{yang2015wikiqa}, TriviaQA~\cite{joshi2017triviaqa}), we did not see direct benefit of SSM. One issue was that some tasks have specialized loss functions, and adapting these to SSM training may require additional care.  
\end{itemize}



\section{Conclusion}
We propose BiGS as a model for pretraining without attention. BiGS makes use of both SSM-based routing and multiplicative gating. Results show that 
SSMs alone perform poorly in a stacked architecture, but gating helps them to generalize. As far as we are aware, this architecture is the first to replicate BERT results without attention.

This work opens up many interesting future questions. We experimented with adapting to longer text, but SSM-based models could be pretrained fully on much longer sequences. Additionally, SSMs have the potential to be significantly faster than attention with further optimization. Finally, we took the first steps in exploring the interesting syntactic properties of SSMs, but it would be interesting to see further probing of how their internal representation leads to these properties.


\section{Ethical Considerations}
Our models are trained using a corpus consisting of existing collections of text from Wikipedia and books. Recent research has uncovered potential societal biases that are embedded within many established corpora. While it is beyond the scope of this paper to delve into these biases in depth, we acknowledge the potential risk that our pre-trained models may inherit these biases. In light of this, we are interested in exploring whether previous research on language bias detection can be applied to BiGS, as part of future work. Additionally, in this paper, we have focused solely on the English corpus, and it would be interesting to investigate how BiGS can contribute to multi-lingual language modeling in the future.


\section{Acknowledgement}
We gratefully acknowledge the support of Google’s TPU
Research Cloud (TRC) program in providing Cloud TPU resources for this research.  AR is supported by NSF CAREER 2037519, NSF 1704834, and a Sloan Fellowship

\bibliography{anthology}
\bibliographystyle{acl_natbib}

\appendix
\section{Appendix}

\subsection{Pre-training Procedure}

\begin{table}[b]
\centering 
\resizebox{\linewidth}{!}{
\begin{tabular}{ccc}
\toprule
Hyperparameter    & BiGS & BERT \\
\midrule
Number of Layers  & 23                   & 24 \\
Hidden size       & 1024                 & 1024 \\
Intermediate size & 3072                 & 4096 \\
Dropout           & 0.1                  & 0.1 \\
Learning Rate Decay & \{Cosine, Linear\} & \{Linear\} \\
Weight Decay      & \{0.05, 0.01\}       & \{0.01\} \\
Learning Rate     & \{2e-4, 4e-4, 6e-4, 8e-4\} & \{2e-4, 4e-4\} \\
Optimizer         & AdamW                & AdamW \\
Adam $\epsilon$   & 1e-6                 & 1e-6 \\
Adam $\beta_1$    & 0.9                  & 0.9 \\
Adam $\beta_2$    & 0.98                 & 0.98 \\
Gradient Clipping & 0.0                  & 0.0 \\
Batch Size        & \{760, 1048, 1136\}  & \{840\} \\
Warmup Proportion & \{1\%\}              & \{2\%\}\\
\bottomrule
\end{tabular}
}
\caption{Hyperparameters used for pretraining BiGS and BERT models}
\label{tab:pretrain}
\end{table}

All models are pretrained using a single cloud TPU-v3. Table~\ref{tab:pretrain} shows hyperparameter configurations that we examine in our pretraining.

BiGS with 512 token length model is trained with 10,000 steps (53,248 tokens per batch) using learning rate 4e-5. 

To compare with LED~\cite{beltagy2020longformer} and BART~\cite{lewis2019bart} in the scroll experiment, we first train a BiGS with 12 layers (119M parameters in total) and 128 maximal sentence length using 500,000 steps and later extend it to 4096 token length with 10k more training steps using learning rate 3e-5.

\subsection{Downstream Tasks}

\begin{table}[t]
\centering 
\resizebox{\linewidth}{!}{
\begin{tabular}{cc}
\toprule
Hyperparameter & GLUE \\
\midrule
Learning Rate     & \{1e-5, 2e-5, 3e-5, 5e-5, 6e-5\} \\
Weight Decay      & \{0.01, 0.1\} \\
Batch Size        & \{16, 32\} \\
Max Epochs        & \{3, 5, 8\} \\
Warmup Proportion & \{0.1\} \\
\bottomrule
\end{tabular}
}
\caption{Hyperparameters used for finetuning our model on GLUE benchmark tasks.}
\label{tab:hyper_glue}
\end{table}

\begin{table}[t]
\centering 
\resizebox{\linewidth}{!}{
\begin{tabular}{ccc}
\toprule
Hyperparameter    & SQuAD & QALT/CNLI \\
\midrule
Learning Rate     & \{4e-5, 6e-5\}    & \{3e-5, 5e-5\} \\
Weight Decay      & \{0, 0.01\} & \{0, 0.01\} \\
Batch Size        & \{32\}      & \{16, 24\} \\
Max Epochs        & \{2\}       & \{5, 8, 10\} \\
Warmup Proportion & \{0.1\}     & \{0.1\} \\
\bottomrule
\end{tabular}
}
\caption{Hyperparameters used for finetuning our model in SQuAD and QALT/CNLI tasks.}
\label{tab:hyper_squad_scrolls}
\end{table}

 All models are finetuned using either a single cloud TPU-v3 or TPU-v2.

\subsubsection{GLUE}
Table~\ref{tab:hyper_glue} shows hyperparameter configurations used to finetune GLUE tasks.

\subsubsection{Other tasks}
Table~\ref{tab:hyper_squad_scrolls} shows hyperparameter configurations used to finetune SQuAD and QALT/CNLI tasks.

\subsection{Annotated CoLA}
\begin{figure*}[ht]
    \centering
    \includegraphics[width=1\linewidth]{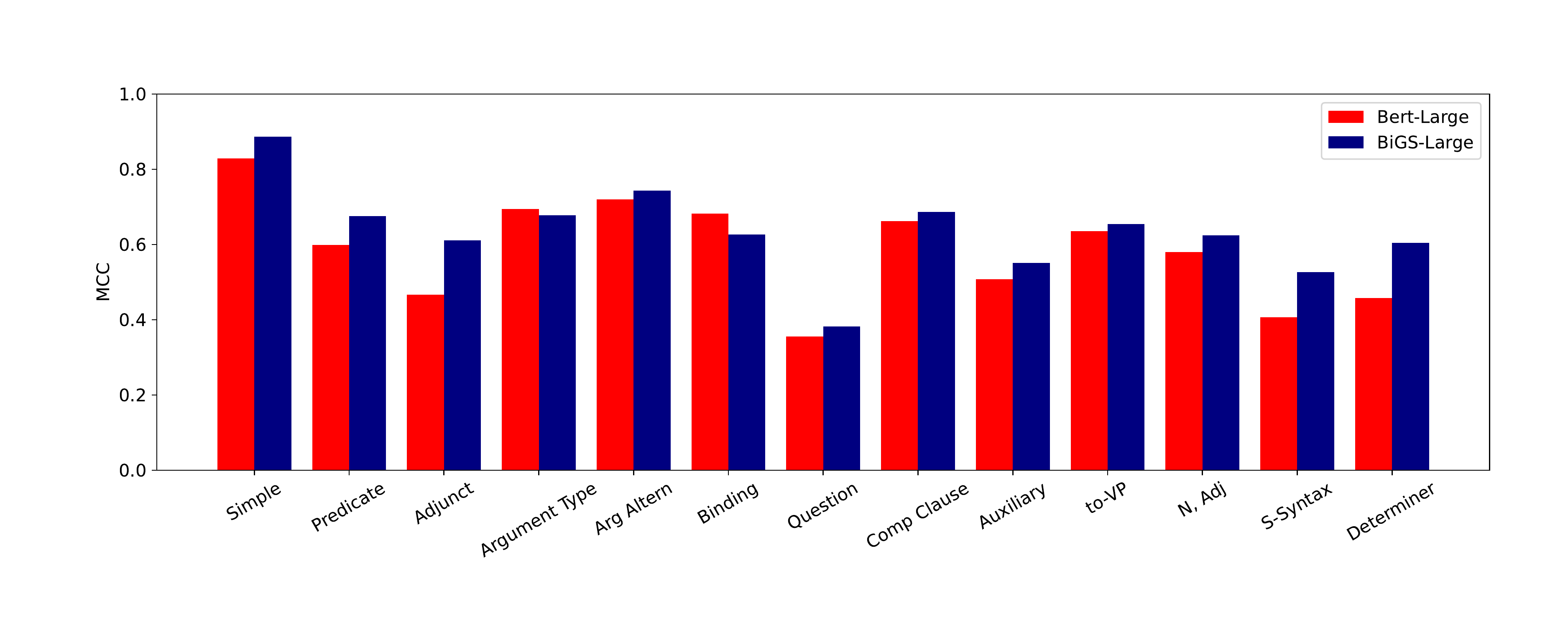}
    \caption{CoLA Results in Different Categories as annotated by \citet{warstadt2019linguistic}. MCC was used to measure the performance. 
    }\label{fig:cola:comparison}
\end{figure*}

The CoLA corpus collection, as described in \citep{warstadt2019neural}, is a vital task within the GLUE benchmark \citep{wang2018glue} for evaluating the acceptability of language models. This corpus has been specifically annotated with 13 different syntactic phenomena in order to more accurately quantify the linguistic knowledge of pre-trained language models (LLMs) \citep{warstadt2019linguistic}. We utilized the annotated instances from this corpus to conduct a detailed analysis of the mistakes made by BiGS and BERT models. Specifically, we used the annotated instances to break down the errors made by these models and understand where they struggle with linguistic knowledge. Results are shown in Figure \ref{fig:cola:comparison}. We discovered that in 9 out of the 13 categories of syntactic phenomena, the BiGS model performed better than the BERT model, and significantly so in two domains. We hypothesize that the inductive bias that BiGS learned during training may have contributed to its superior performance in understanding these syntactic phenomena. It is likely that the specific inductive biases encoded in the BiGS model enabled it to better comprehend the nuances of these syntactic phenomena, leading to its improved performance.

\end{document}